\pdfoutput=1

\documentclass[10pt]{article}

\usepackage{acl}

\usepackage{times}
\usepackage{latexsym}
\usepackage[T1]{fontenc}
\usepackage[utf8]{inputenc}
\usepackage{microtype}
\usepackage{inconsolata}
\usepackage{graphicx}
\usepackage{booktabs}
\usepackage{natbib}
\usepackage{pdfpages}
\usepackage{pifont} 

\title{Bi-Fact: A Bidirectional Factorization-based Evaluation 
\\ of Intent Extraction from UI Trajectories}

\author{
    Sapir Caduri \textsuperscript{1}
    \quad Anatoly Efros \textsuperscript{1}
    \quad Noam Kahlon \textsuperscript{1}
    \quad Danielle Cohen \textsuperscript{1} \\
    \quad \textbf{Yoni Halpern \textsuperscript{1}}
     \quad \textbf{Ido Dagan \textsuperscript{1,2}} \\
     \textsuperscript{1}Google Research \quad \textsuperscript{2}Bar-Ilan University \\
     \texttt{\small\{sapir,talef,kahlonn,daniellecn,yhalpern,idodagan\}@google.com} \\
}

\begin{document}
\maketitle
\begin{abstract}
Evaluating user intent extraction from GUI interactions demands accurate, fine-grained metrics. This paper introduces Bi-Fact, a novel method that decomposes intents into atomic facts and performs bidirectional comparisons with reference intents in order to assess precision and recall.  Experiments demonstrate Bi-Fact's superior correlation with human judgments compared to existing metrics, establishing a more robust automated evaluation framework for UI-driven intent understanding.

\end{abstract}

\section{Introduction}

Understanding user intent from interactions with graphical user interfaces (GUIs) is an area of growing interest. This involves translating sequences of UI events, such as clicks, scrolls, and text inputs, into a high-level description of the user's desired outcome. This extracted intent is crucial for various downstream applications, including personalized user experiences, and proactive suggestions.

Notably, these applications depend on the accuracy of individual extracted facts. For instance, consider the following gold intent ``book a business-class flight to Paris'' where the model prediction was ``book a flight to Paris''. Although the prediction is not fully correct, some of the facts within the gold intent were correctly identified. Obtaining them would allow downstream models, for example, to generate follow-up actions for the user and learn useful personal profile items relating to the users activity. Therefore, a robust evaluation framework is needed to assess the accuracy of intent predictions at a granular, fact-based level.

Recent research has begun to address intent extraction from UI interactions, developing various models for the task \cite{berkovitch2024identifying, sun2024genesis, huang2024automatic, martinez2024screenshot, zhang2024summact, yang2024aria}. 
In this task setting, a model receives as input a trajectory of a GUI session, consisting of screenshots and user actions, and generates a predicted natural-language intent description, which is supposed to fully and precisely describe the user intents which were fulfilled during the session.
However, evaluating these extracted intents remains a hurdle. Existing metrics struggle to capture the nuances of UI-driven intents, especially regarding fine-grained factual accuracy and completeness.
Traditional text similarity measures like BLEU \cite{papineni2002bleu} and ROUGE \cite{lin2004rouge} rely on lexical overlap and are ill-suited for capturing semantic equivalence in UI contexts. On the other hand, sentence-level semantic similarity metrics and inference-based methods like classifying textual-entailment with Natural Language Inference (NLI) models \cite{ido2006Pascal, bowman2015large} also fall short. While full-intent level NLI can assess overall entailment between two intents, it does not capture the fine-grained factual discrepancies crucial for downstream tasks. 

For example, a detailed request for a trip booking on Monday and a similar request for that trip on Tuesday are related but do not strictly entail or contradict each other, potentially leading NLI models to miss the pivotal differences in dates.  
Fact-level evaluation methods, such as Pyramid \cite{nenkova2004evaluating} and its automated variants \cite{gao2019automated,yang2016peak,harnly2005automation},, or FactScore \cite{min2023factscore}, offer a more granular approach by comparing the generated and reference texts at a granular level of factual claims, which are extracted from the respective text.
However, while Pyramid is a recall-oriented measure and FactScore is precision-oriented, evaluating intent understanding requires considering both the recall (are all user-intended facts captured?) and precision (are all extracted facts correct and relevant?) of the extracted factual information relative to the gold intent. 

In this work, we propose Bi-Fact, a novel fact-level evaluation method specifically designed for assessing intents extracted from UI interactions, enabling a more detailed and accurate evaluation than prior metrics utilized for this setting. The core idea is to first decompose both predicted and gold intents into their constituent facts, and then to test for the support for each fact in the counterpart intent. We demonstrate that Bi-Fact exhibits significantly high correlation with human judgments of fact-level intent equivalence, providing a robust and informative evaluation framework for assessing intent extraction in UI interaction scenarios. 

\begin{figure*}[t]
    \centering
    \includegraphics[width=\linewidth]{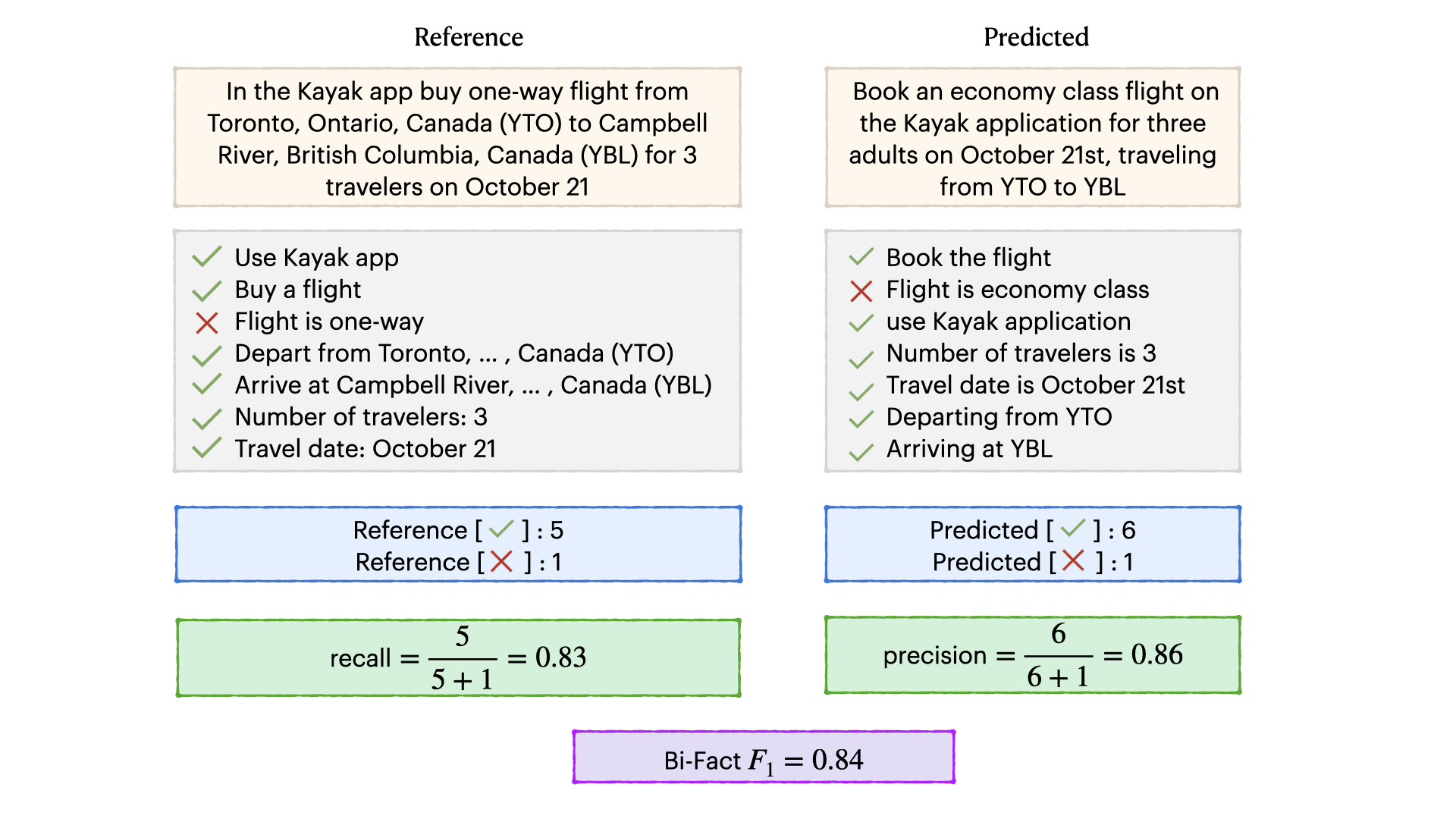}
    \caption{Example of Bi-Fact Evaluation Process: Comparing a Gold (reference) intent with a Predicted (model-generated) intent. Each intent is decomposed into atomic facts. Facts are compared  bidirectionally for their support by the other text, as represented with checkmarks (\ding{51}) for implied facts and crosses (\ding{55}) for missed facts. Recall, precision, and F1 scores are computed from the bidirectional comparison.}
    \label{fig:autofact_figure}
\end{figure*}

\section{The Bi-Fact method of Evaluating Intent Extraction}


\subsection{Approach}
At a high-level, given a gold intent and a predicted intent, we would like to decompose both intents into atomic facts, representing single, indivisible pieces of information. 
For example, the intent ``book a flight to Paris for a weekend business trip'', might be decomposed into facts like ``book a flight'', ``destination is Paris'', ``trip type is business'', and ``duration is weekend''. 
We then perform a bidirectional evaluation: checking whether each fact in the predicted intent is implied (entailed) from the gold intent (precision), and conversely, whether each fact in the gold intent can is implied from the predicted intent (recall). The computation flow is summarized in Figure~\ref{fig:autofact_figure}.

\subsection{Automatic Evaluation with Bi-Fact}

We automate the Bi-Fact evaluation protocol utilizing a large language model (LLM). The methodology consists of the following two primary parts, with the assessment step further divided into three reasoning stages:

\textbf{Part 1: Factorizing the Gold Intents:} Initially, given a dataset with gold intents (along with their corresponding trajectories), the gold intents are factorized into atomic facts. Following this stage, the factual decomposition of the gold intents is fixed, providing a stable basis for subsequent comparative evaluations of the predictions made by various models. The prompt used for this factorization is detailed in Appendix~\ref{appendix:prompts} (Figure ~\ref{fig:bifact_factorize_prompt}).

\textbf{Part 2: Factual Coverage Assessment:} This part is performed when evaluating the predicted intents generated by an intent extraction model. It consists of three stages, first decomposing the predicted intent into a set of individual facts, and then assessing the recall and precision of the predicted intent relative to the gold intent. This phase is performed in a single prompt (Figure~\ref{fig:assessment_prompt_p1}).

\begin{enumerate}
    \item \textbf{Stage 1: Factorizing the predicted intent:} The LLM first decomposes the predicted intent into its atomic facts. As an additional input, the LLM receives the decomposition of the gold intent (as fixed in Part 1), and is instructed to decompose the predicted intent in accordance with the given decomposition of the gold intent, to increase their consistency for comparison.
    \item \textbf{Stage 2: Recall assessment:} The LLM examines each gold fact (from the decomposed gold intent, obtained in Part 1), and labels it as either implied (entailed) or not-implied from the predicted intent, where the latter is considered as a whole. Recall is then measured as the proportion of implied facts among the gold facts.
    \item \textbf{Stage 3: Assessing the predicted facts:} The LLM examines each predicted fact (from the decomposed predicted intent, obtained in Stage 1), and labels it as either implied (entailed) or not-implied from the gold intent, where the latter is considered as a whole. Precision is then measured as the proportion of implied facts among the predicted facts.
\end{enumerate}

Bi-Fact's strength lies in its atomic-level operation. This granular approach breaks down complex sentences for fact-level comparison, thus
supporting a comprehensive assessment of both precision and recall, while providing precise information regarding which facts are supported and which are not, in both the gold and predicted intents.




\section{Evaluating Bi-Fact}
\subsection{Datasets}
\label{sec:datasets}

We utilize two datasets to evaluate the effectiveness of our proposed Bi-Fact metric. These datasets include human assessments of predicted intents, in comparison to given gold intents. Thus, these datasets allow measuring the agreement between these human intent evaluations and automatic evaluation measures, which are applied to the same gold and predicted intent pairs.

\paragraph {\textbf Intent-Match Data:} To evaluate our proposed metric, we utilize the dataset from \cite{berkovitch2024identifying}, which provides manually annotated intent data. The dataset comprises 100 examples sampled from the test sets of Mind2Web \cite{deng2024mind2web} and AitW \cite{rawles2024androidinthewild}. Each example includes a human-generated intent describing the UI trajectory and model-generated intent predictions from Gemini and GPT. This results in a total of 400 gold-predicted intent pairs. Each pair was manually annotated with a binary judgement, of whether both intents indicate the same task, in the sense that two trajectories, fulfilling the gold and predicted intent, respectively, would achieve the same outcome. 
Accordingly, this dataset is suitable to assess automatic evaluation measures that yield a binary judgement of whether a predicted intent, as a whole, is semantically equivalent to the gold intent, with respect to the task it defines.
We allocated 10\% of this data as development set to calibrate the thresholds of the different intent evaluation methods, and the remaining 90\% were used as a test set. This dataset facilitates a comprehensive evaluation of intent comparison across manual annotations from multiple datasets and model-generated predictions from various sources, making it suitable for assessing the performance of our proposed intent evaluation metric.
    
\paragraph{\textbf Fact-Level Data:} To validate the correctness of our automatic method at the granular fact-level assessment, we conducted a thorough manual factorization on a small dataset of 36 intent pairs. We then manually assessed the entailment of the decomposed intent facts across the gold and predicted intents in each instance, yielding the respective precision and recall scores and the corresponding F1 scores based on these annotations. In Section~\ref{sec:fact_level_results}, we demonstrate a high correlation between our automatic method and these manual annotations.

\subsection{Comparison to other evaluation approaches on binary Intent-match }

For the binary evaluation judgment (whether a predicted intent is equivalent to a gold intent, as a whole), we compare \textit{Bi-Fact} against standard text similarity metrics, including BLEU \cite{papineni2002bleu}, ROUGE \cite{lin2004rouge}, METEOR \cite{banerjee2005meteor}, T5 cosine similarity~\citep{ni2021sentence}, and NLI-based methods \cite{honovich-etal-2022-true-evaluating}. We binarized the continuous scores using thresholds that yielded the best performance in terms of F1 score on the development set (testing 30 equally spaced increments between 0.01 and 1.0). For the NLI-based method, we used a bidirectional approach, taking the average of the NLI scores for both directions (premise vs. hypothesis and hypothesis vs. premise). We also include the Automatic Rater (AutoRater) introduced in \cite{berkovitch2024identifying}. For methods that yield a numeric score (rather than a binary judgement), we tune a hyper-parameter of a score threshold for positive judgement (match vs. non-match) over the development set of the Intent-Match data.
To assess the effectiveness of these metrics, we evaluate the correlation between each of these scores and human judgments on the intent-match data using recall, precision, F1 score, and kappa. The results are in Table~\ref{tab:evaluation_metrics}. Bi-Fact notably achieves the highest F1 score (0.722) and Kappa coefficient (0.508) among all tested metrics, indicating better agreement with human judgments. The results highlight the limitations of purely lexical overlap based metrics for capturing the nuances of intent in UI interactions. 

\begin{table}[ht]
    \centering
    
    \begin{tabular}{lcccc}
        \hline
        \textbf{Metric} & \textbf{Precision} & \textbf{Recall} & \textbf{F1} & \textbf{Kappa} \\
        \hline
        BLEU       & 0.529     & 0.787  & 0.632   & 0.301 \\
        ROUGE-1    & 0.516     & 0.890  & 0.653    & 0.304 \\
        ROUGE-2    & 0.513     & 0.732  & 0.603    & 0.257 \\
        ROUGE-L    & 0.511     & 0.835  & 0.634    & 0.281 \\
        METEOR     & 0.536     & 0.811  & 0.646    & 0.323 \\
        NLI        & 0.895     & 0.470   & 0.614  & 0.467 \\
        T5 sim       & 0.963    & 0.513  & 0.669     & 0.316 \\
        AutoRater  & 0.909     & 0.382  & 0.538    & 0.395 \\ 
        Bi-Fact   & 0.658     & 0.799  & \textbf{0.722}   & \textbf{0.508} \\ 

        \hline
    \end{tabular}
    \caption{Comparing different text comparison metrics' effectiveness at evaluating binary intent match. Bidirectional comparisons are used, and continuous scores are binarized using a threshold.}
    \label{tab:evaluation_metrics}
\end{table}

\subsection{Comparison using Fact-level annotations}
\label{sec:fact_level_results}

We further validated Bi-Fact by analyzing its correlation with manual fact-level annotations. This analysis was conducted on the Manual Fact-Level dataset described in Section~\ref{sec:datasets}, providing insights into Bi-Fact's ability to capture fine-grained factual information, which is highly crucial and informative for downstream tasks, as discussed in the Introduction. Bi-Fact's F1 score demonstrates a strong correlation with the human fact-level annotations, achieving a Pearson correlation of 0.781 (p < 0.001). This substantially outperformed the second-best method, NLI, which reached only 0.517 (p = 0.001).

\section{Discussion}

The superior performance of Bi-Fact can be attributed to its ability to decompose intents into atomic facts, enabling it to capture deeper semantic and functional similarities between intents, which are crucial for UI-based applications. The high correlation with manual fact-level annotations further demonstrates Bi-Fact's effectiveness in accurately assessing the factual alignment between intents.

\bibliography{bibliography}

\begin{thebibliography}{20}
\providecommand{\natexlab}[1]{#1}

\bibitem[{Banerjee and Lavie(2005)}]{banerjee2005meteor}
Satanjeev Banerjee and Alon Lavie. 2005.
\newblock Meteor: An automatic metric for mt evaluation with improved
  correlation with human judgments.
\newblock In \emph{Proceedings of the acl workshop on intrinsic and extrinsic
  evaluation measures for machine translation and/or summarization}, pages
  65--72.

\bibitem[{Berkovitch et~al.(2024)Berkovitch, Caduri, Kahlon, Efros, Caciularu,
  and Dagan}]{berkovitch2024identifying}
Omri Berkovitch, Sapir Caduri, Noam Kahlon, Anatoly Efros, Avi Caciularu, and
  Ido Dagan. 2024.
\newblock Identifying user goals from ui trajectories.
\newblock \emph{arXiv preprint arXiv:2406.14314}.

\bibitem[{Bowman et~al.(2015)Bowman, Angeli, Potts, and
  Manning}]{bowman2015large}
Samuel~R Bowman, Gabor Angeli, Christopher Potts, and Christopher~D Manning.
  2015.
\newblock A large annotated corpus for learning natural language inference.
\newblock \emph{arXiv preprint arXiv:1508.05326}.

\bibitem[{Dagan et~al.(2006)Dagan, Glickman, and Magnini}]{ido2006Pascal}
Ido Dagan, Oren Glickman, and Bernardo Magnini. 2006.
\newblock The pascal recognising textual entailment challenge.
\newblock In \emph{Machine Learning Challenges. Evaluating Predictive
  Uncertainty, Visual Object Classification, and Recognising Tectual
  Entailment}, pages 177--190, Berlin, Heidelberg. Springer Berlin Heidelberg.

\bibitem[{Deng et~al.(2024)Deng, Gu, Zheng, Chen, Stevens, Wang, Sun, and
  Su}]{deng2024mind2web}
Xiang Deng, Yu~Gu, Boyuan Zheng, Shijie Chen, Sam Stevens, Boshi Wang, Huan
  Sun, and Yu~Su. 2024.
\newblock Mind2web: Towards a generalist agent for the web.
\newblock \emph{Advances in Neural Information Processing Systems}, 36.

\bibitem[{Gao et~al.(2019)Gao, Sun, and Passonneau}]{gao2019automated}
Yanjun Gao, Chen Sun, and Rebecca~J Passonneau. 2019.
\newblock Automated pyramid summarization evaluation.
\newblock In \emph{Proceedings of the 23rd Conference on Computational Natural
  Language Learning (CoNLL)}.

\bibitem[{Harnly et~al.(2005)Harnly, Nenkova, Passonneau, and
  Rambow}]{harnly2005automation}
Aaron Harnly, Ani Nenkova, Rebecca Passonneau, and Owen Rambow. 2005.
\newblock Automation of summary evaluation by the pyramid method.
\newblock In \emph{Recent Advances in Natural Language Processing (RANLP)},
  pages 226--232.

\bibitem[{Honovich et~al.(2022)Honovich, Aharoni, Herzig, Taitelbaum,
  Kukliansy, Cohen, Scialom, Szpektor, Hassidim, and
  Matias}]{honovich-etal-2022-true-evaluating}
Or~Honovich, Roee Aharoni, Jonathan Herzig, Hagai Taitelbaum, Doron Kukliansy,
  Vered Cohen, Thomas Scialom, Idan Szpektor, Avinatan Hassidim, and Yossi
  Matias. 2022.
\newblock \href {https://doi.org/10.18653/v1/2022.naacl-main.287} {{TRUE}:
  Re-evaluating factual consistency evaluation}.
\newblock In \emph{Proceedings of the 2022 Conference of the North American
  Chapter of the Association for Computational Linguistics: Human Language
  Technologies}, pages 3905--3920, Seattle, United States. Association for
  Computational Linguistics.

\bibitem[{Huang et~al.(2024)Huang, Li, Li, and Li}]{huang2024automatic}
Forrest Huang, Gang Li, Tao Li, and Yang Li. 2024.
\newblock Automatic macro mining from interaction traces at scale.
\newblock In \emph{Proceedings of the CHI Conference on Human Factors in
  Computing Systems}, pages 1--16.

\bibitem[{Lin(2004)}]{lin2004rouge}
Chin-Yew Lin. 2004.
\newblock Rouge: A package for automatic evaluation of summaries.
\newblock In \emph{Text Summarization Branches Out: Proceedings of the ACL
  Workshop}, pages 74--81.

\bibitem[{Mart{\'\i}nez-Rojas et~al.(2024)Mart{\'\i}nez-Rojas,
  Jim{\'e}nez-Ram{\'\i}rez, Enr{\'\i}quez, and
  Reijers}]{martinez2024screenshot}
Antonio Mart{\'\i}nez-Rojas, A~Jim{\'e}nez-Ram{\'\i}rez, Jos{\'e}~Gonzalez
  Enr{\'\i}quez, and Hajo~A Reijers. 2024.
\newblock A screenshot-based task mining framework for disclosing the drivers
  behind variable human actions.
\newblock \emph{Information Systems}, 121:102340.

\bibitem[{Min et~al.(2023)Min, Krishna, Lyu, Lewis, Yih, Koh, Iyyer,
  Zettlemoyer, and Hajishirzi}]{min2023factscore}
Sewon Min, Kalpesh Krishna, Xinxi Lyu, Mike Lewis, Wen-tau Yih, Pang~Wei Koh,
  Mohit Iyyer, Luke Zettlemoyer, and Hannaneh Hajishirzi. 2023.
\newblock Factscore: Fine-grained atomic evaluation of factual precision in
  long form text generation.
\newblock \emph{arXiv preprint arXiv:2305.14251}.

\bibitem[{Nenkova and Passonneau(2004)}]{nenkova2004evaluating}
Ani Nenkova and Rebecca~J Passonneau. 2004.
\newblock Evaluating content selection in summarization: The pyramid method.
\newblock In \emph{Proceedings of the human language technology conference of
  the north american chapter of the association for computational linguistics:
  Hlt-naacl 2004}, pages 145--152.

\bibitem[{Ni et~al.(2021)Ni, Abrego, Constant, Ma, Hall, Cer, and
  Yang}]{ni2021sentence}
Jianmo Ni, Gustavo~Hernandez Abrego, Noah Constant, Ji~Ma, Keith~B Hall, Daniel
  Cer, and Yinfei Yang. 2021.
\newblock Sentence-t5: Scalable sentence encoders from pre-trained text-to-text
  models.
\newblock \emph{arXiv preprint arXiv:2108.08877}.

\bibitem[{Papineni et~al.(2002)Papineni, Roukos, Ward, and
  Zhu}]{papineni2002bleu}
Kishore Papineni, Salim Roukos, Todd Ward, and Wei-Jing Zhu. 2002.
\newblock Bleu: a method for automatic evaluation of machine translation.
\newblock In \emph{Proceedings of the 40th annual meeting of the Association
  for Computational Linguistics}, pages 311--318.

\bibitem[{Rawles et~al.(2024)Rawles, Li, Rodriguez, Riva, and
  Lillicrap}]{rawles2024androidinthewild}
Christopher Rawles, Alice Li, Daniel Rodriguez, Oriana Riva, and Timothy
  Lillicrap. 2024.
\newblock Androidinthewild: A large-scale dataset for android device control.
\newblock \emph{Advances in Neural Information Processing Systems}, 36.

\bibitem[{Sun et~al.(2024)Sun, Cheng, Ding, Jin, Wang, Xu, Wu, Jia, Chen, Liu
  et~al.}]{sun2024genesis}
Qiushi Sun, Kanzhi Cheng, Zichen Ding, Chuanyang Jin, Yian Wang, Fangzhi Xu,
  Zhenyu Wu, Chengyou Jia, Liheng Chen, Zhoumianze Liu, et~al. 2024.
\newblock Os-genesis: Automating gui agent trajectory construction via reverse
  task synthesis.
\newblock \emph{arXiv preprint arXiv:2412.19723}.

\bibitem[{Yang et~al.(2016)Yang, Passonneau, and De~Melo}]{yang2016peak}
Qian Yang, Rebecca Passonneau, and Gerard De~Melo. 2016.
\newblock Peak: Pyramid evaluation via automated knowledge extraction.
\newblock In \emph{Proceedings of the AAAI conference on artificial
  intelligence}, volume~30.

\bibitem[{Yang et~al.(2024)Yang, Wang, Li, Luo, Chen, Huang, and
  Li}]{yang2024aria}
Yuhao Yang, Yue Wang, Dongxu Li, Ziyang Luo, Bei Chen, Chao Huang, and Junnan
  Li. 2024.
\newblock Aria-ui: Visual grounding for gui instructions.
\newblock \emph{arXiv preprint arXiv:2412.16256}.

\bibitem[{Zhang et~al.(2024)Zhang, Ahmed, Hu, and Bulling}]{zhang2024summact}
Guanhua Zhang, Mohamed Ahmed, Zhiming Hu, and Andreas Bulling. 2024.
\newblock Summact: Uncovering user intentions through interactive behaviour
  summarisation.
\newblock \emph{arXiv preprint arXiv:2410.08356}.

\end{thebibliography}

\onecolumn 

\appendix

\section{Prompts}
\label{appendix:prompts}

\begin{figure}[h!] 
  
  \centering 
  \includegraphics[width=\textwidth]{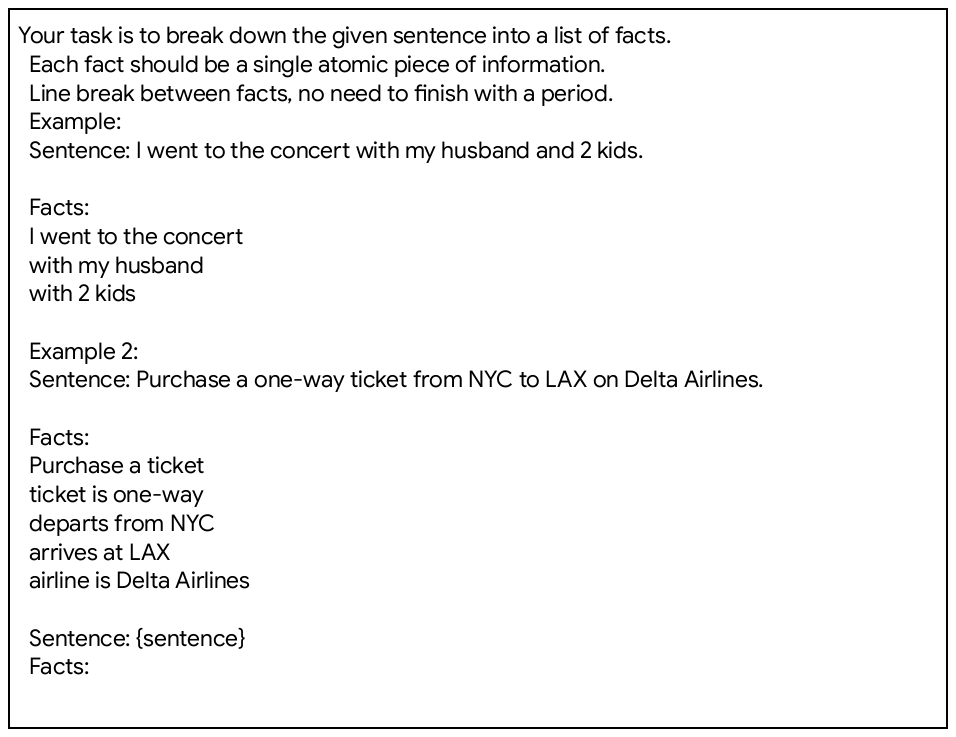}  
  \caption{Example prompt for the fact decomposition task} 
  \label{fig:bifact_factorize_prompt} 
\end{figure}

\newpage 

\begin{figure}[htp] 
  \centering 
  \includegraphics[width=\textwidth]{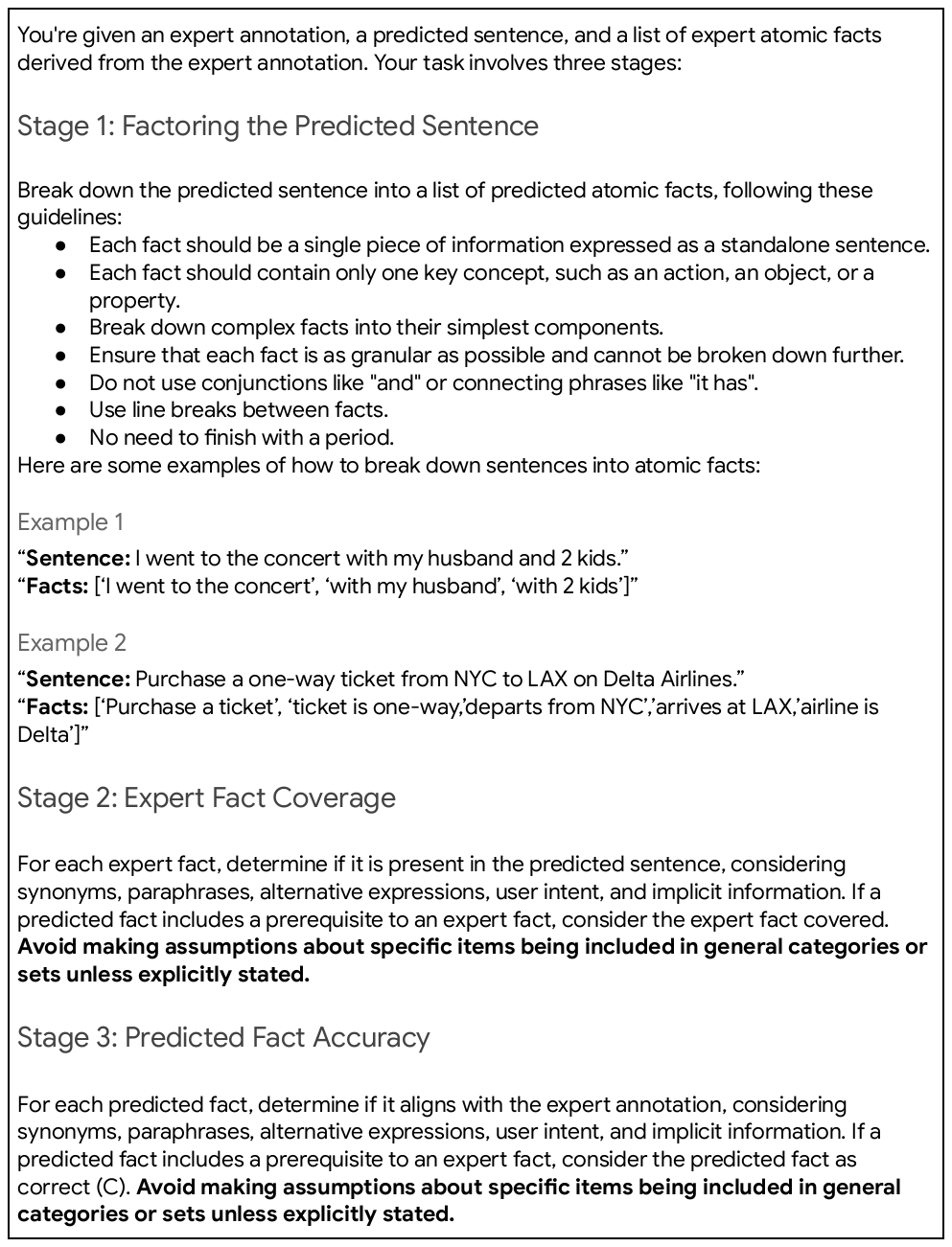}  
  \caption{Assessment prompt part 1 - 3 step instructions} 
  \label{fig:assessment_prompt_p1} 
\end{figure}

\begin{figure}[ht] 
  \centering 
  \includegraphics[width=\textwidth]{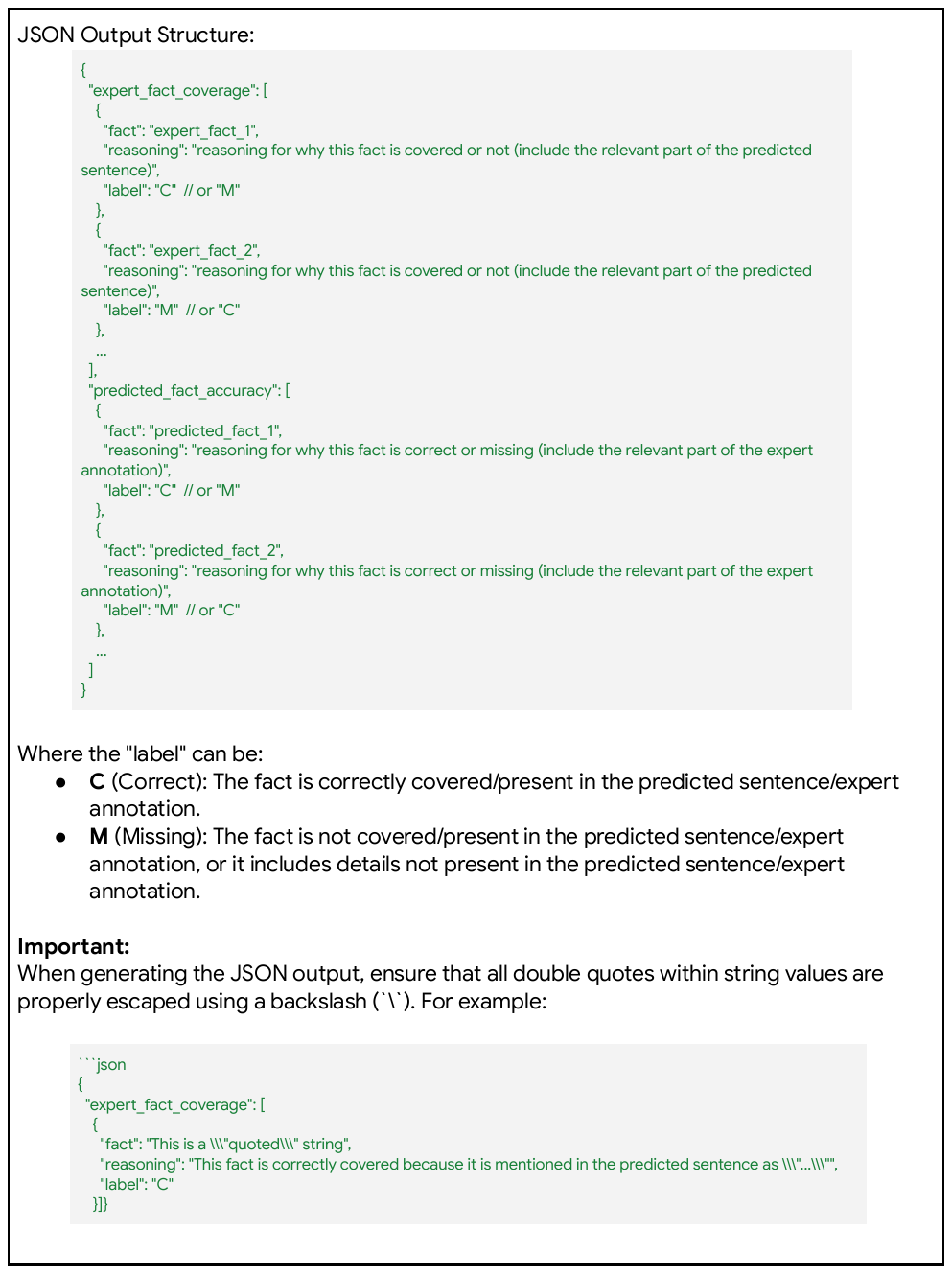}  
  \caption{Assessment prompt part 2 - Output Structure} 
  \label{fig:assessment_prompt_p2} 
\end{figure}

\begin{figure}[ht] 
    \centering 
    \includegraphics[width=\textwidth]{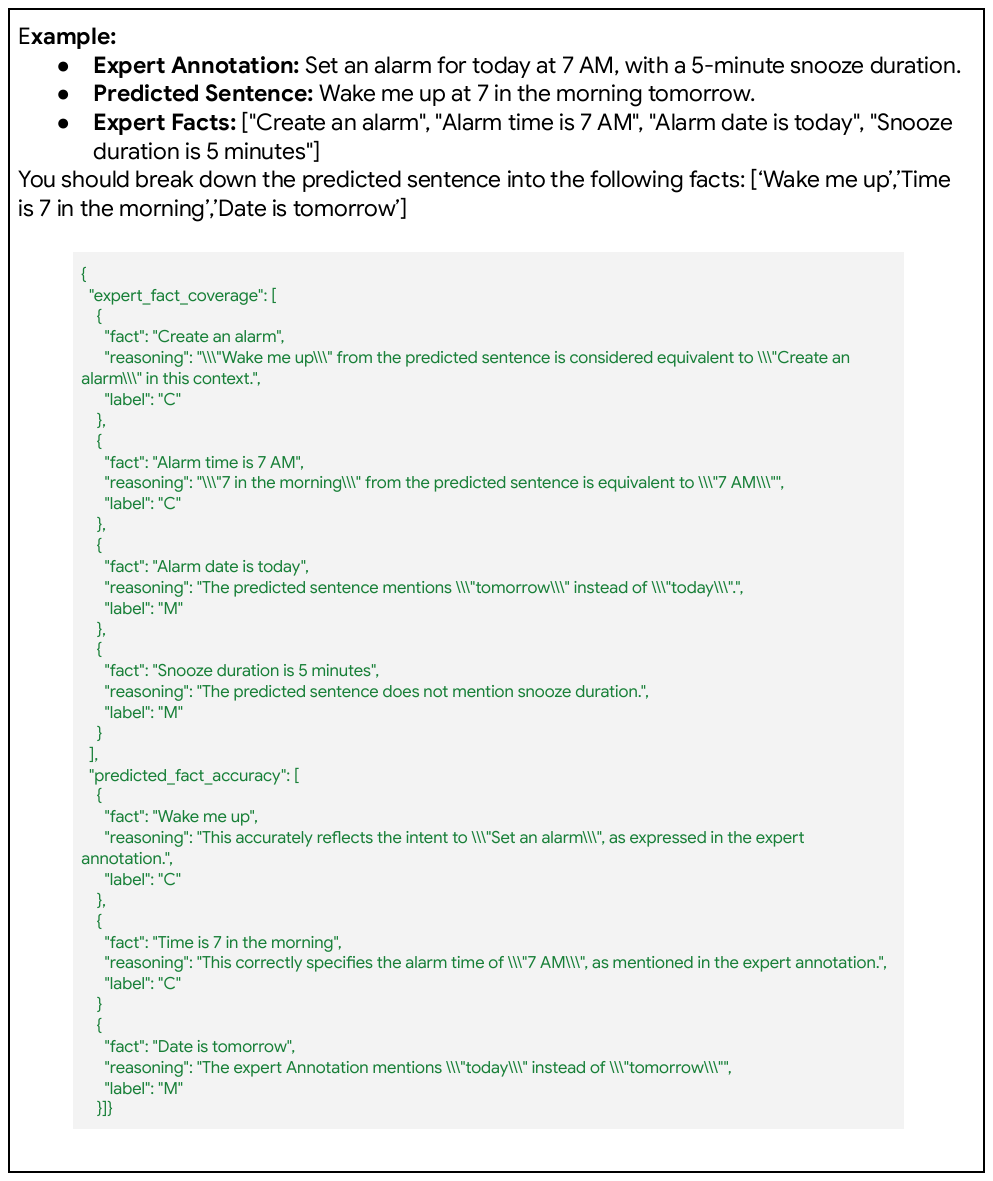}  
    \caption{Assessment prompt part 3 - An Example (one-shot)} 
    \label{fig:assessment_prompt_p3} 
\end{figure}

\end{document}